\documentclass[conference]{IEEEtran}
\IEEEoverridecommandlockouts
\usepackage{cite}
\usepackage{amsmath,amssymb,amsfonts}
\usepackage{amsthm}
\usepackage{graphicx}
\usepackage{textcomp}
\usepackage{xcolor}

\usepackage{algorithm}
\usepackage[noend]{algpseudocode}
\usepackage{booktabs}
\usepackage[font=small]{caption}
\usepackage{caption, subcaption}

\makeatletter
\let\blx@rerun@biber\relax
\makeatother

\def\BibTeX{{\rm B\kern-.05em{\sc i\kern-.025em b}\kern-.08em
		T\kern-.1667em\lower.7ex\hbox{E}\kern-.125emX}}

\begin{document}
	
	\title{Analysis of Evolutionary Behavior in Self-Learning Media Search Engines}
	
	
	\author{\IEEEauthorblockN{Nikki Lijing Kuang\IEEEauthorrefmark{1},
			Clement H.C. Leung\IEEEauthorrefmark{2}\IEEEauthorrefmark{3}}
		\IEEEauthorblockA{\IEEEauthorrefmark{1}
			\textit{Department of Computer Science and Engineering},
			\textit{University of California San Diego}, La Jolla, USA\\
			\IEEEauthorrefmark{2}
			\textit{School of Science and Engineering},
			\textit{The Chinese University of Hong Kong}, Shenzhen, China\\
			\IEEEauthorrefmark{3}\textit{Shenzhen Research Institute of Big Data}, Shenzhen, China\\
			l1kuang@ucsd.edu, clementleung@cuhk.edu.cn
	}}
	
	\maketitle
	
	\begin{abstract}
		The diversity of intrinsic qualities of multimedia entities tends to impede their effective retrieval. In a Self-Learning Search Engine architecture, the subtle nuances of human perceptions and deep knowledge are taught and captured through unsupervised reinforcement learning, where the degree of reinforcement may be suitably calibrated. Such architectural paradigm enables indexes to evolve naturally while accommodating the dynamic changes of user interests. It operates by continuously constructing indexes over time, while injecting progressive improvement in search performance. For search operations to be effective, convergence of index learning is of crucial importance to ensure efficiency and robustness. In this paper, we develop a Self-Learning Search Engine architecture based on reinforcement learning using a Markov Decision Process framework. The balance between exploration and exploitation is achieved through evolutionary exploration Strategies. The evolutionary index learning behavior is then studied and formulated using stochastic analysis. Experimental results are presented which corroborate the steady convergence of the index evolution mechanism.
	\end{abstract}
	
	
	\section{Introduction and Related Works}
	With the explosive growth of multimedia information in contemporary social networks, effective retrieval of a wide variety of digital media has drawn increasing attention in both business and research realms \cite{matsui2017sketch, song2018quantization}. Current prevailing search methods, such as PageRank, tf-idf, and BGSA \cite{rashedi2010bgsa}, has achieved impressive performance in classic text retrieval \cite{lofgren2016personalized, aggarwal2018information}. Nevertheless, the emergence of distinct genres of high-dimensional multimedia entities \cite{zhao2017real, zhu2016deep}, such as avant-garde 3D models and culture artifacts, has introduced higher complexity and unfulfilled challenges to multimedia and cross-media search \cite{gao2017learning, pouyanfar2018multimedia, bello2016social}. 
	
	One of the basic and long-standing issues in different forms of multimedia information retrieval is to address the semantic gap between low-level features and high-level semantics \cite{over2004multimedia, azzam2004implicit, stevenson2005comparative}. To resolve the issue, in Content-Based Image Retrieval (CBIR) systems, general supervised learning algorithms with deep neural networks are adopted for deep knowledge learning \cite{iakovidou2019composite, bai2018optimization}. In the relatively young and challenging field of Music Information Retrieval (MIR), the wide adoption of practical musical recommendation systems induces a research shift to contextual feature categories \cite{jin2019contextplay, volokhin2018towards, wang2018learning}, which could be associated with audio-based techniques for enhancement \cite{oramas2017sound, gupta2018explicit}. On the other hand, in the more sophisticated Video Retrieval where visual attributes, audio features, and narration text content are coupled, with standard protocol disappointingly absent \cite{pardhi2016performance, hong2018cbvmr}, partly due to the lack of high-quality training dataset and supporting information for queries \cite{cheng2016information}. Popular techniques such as deep learning and agents networks are often deployed to effect improvements in performance \cite{yu2018joint, dong2018deep, zhang2008topological, song2018self}, and existing methods sometimes involve fusing distinct categories of information together through feature learning \cite{lokovc2018influential}. Likewise, research in the newly-emerged cross-media retrieval attempts to project heterogeneous features into a common latent feature space to facilitate similarity computation \cite{yao2019discrete, dong2018cross, deng2018triplet, wei2016modality}. However, such tradeoff generally fails to deliver the best performance. 
	
	The lack of an efficient generalization method that is independent of media modalities imposes an imperative need for a novel architecture to retrieve diversified forms of data. Unlike most of the above methods that rely on content-based analysis, indexing technique empowered by relevance feedback (RF) \cite{leung2012intelligent} has shown to be promising for significant performance improvement in multimedia and cross-media search engines \cite{ cobarzan2017interactive}. The search engine architectural paradigm proposed in \cite{leung2012intelligent} captures the knowledge of human perceptions by adopting an implicit RF technique to empower the system for evolution; however, rigorous performance analysis has not been performed.
	
	Meanwhile, index learning has generated considerable interests in multimedia information retrieval \cite{awad2017video, datta2017multimodal, mohamed2014fast, amato2016large}, where indexing is enabled by utilizing available textual metadata or RF information to improve search precision. In \cite{datta2017multimodal}, Datta et al. use indexing and matching score calculation to expand textual queries on both image search and the corresponding text retrieval. By developing indexing and searching algorithms for Metric Permutation Table, Mohamed et. al. achieves efficient retrieval in large-scale multimedia databases \cite{mohamed2014fast}. To support efficient CBIR in large image sets, Amato \cite{amato2016large} makes use of Deep Convolutional Neural Networks.
	
	In this paper, we develop a novel Self-Learning Search Engine (SLSE) Architecture based on Reinforcement Learning (RL) that continuously learns to adjust its actions to meet changing demands. RL is a natural paradigm to be widely adopted for addressing the problem of sequential decision making. It is concerned with the task of how an intelligent agent takes optimal actions in interaction with an environment through a trial-and-error learning process, with the ultimate goal to maximize the long-term rewards \cite{sutton2018reinforcement}. Since the development of AlphaGo \cite{silver2016mastering} and AlphaGo Zero \cite{silver2017mastering}, RL has been widely adopted in different contexts \cite{luong2019applications, hu2018reinforcement, zheng2018drn, lee2018interactive}. By learning from useful evaluative information provided by users it is shown to be a viable solution for web recommendation \cite{zheng2018drn} and information retrieval systems \cite{hu2018reinforcement, rosset2018optimizing} to yield satisfactory recommendations and search results. 
	
	The main contributions of this paper are as follows:
	\begin{itemize}
		\item We introduce a novel reinforcement learning based Self-Learning Search Engine (SLSE) Architecture for multimedia search and multimedia data retrieval. The presented SLSE paradigm is capable of dynamically adjusting its inner workings to adapt to the latest user interests by learning online from the collective feedback behavior of users.
		\item We propose a dynamic indexing technique for SLSE architecture that is able to dynamically build up index tuples according to relevance and destruct historical ones to handle exceptions along with time.
		\item To prevent local optimum and to enable proactive exploration, we propose evolutionary exploration strategies to retrieve search results, which balance the commonly occurring exploitation and exploration problem in reinforcement learning.
		\item We study the problem of learning convergence by undertaking rigorous analysis from the theoretical perspective. Closed-form solutions are obtained, which support the viability of SLSE architecture. The evaluations and experiments also corroborate the theoretical results. 
	\end{itemize}
	
	The rest of the paper is organized as follows: Section II provides the essential background information of techniques involved in SLSE, followed by Section III that describes the architecture of SLSE and formulates it as a Markov Decision Process. Specifically, we elaborate on the problem of learning convergence that concerns the performance of SLSE. Section IV then analyzes the learning convergence behavior theoretically to provide systematic and closed-form evaluation. Experimental results are given in Section V, followed by the conclusions in Section VI.
	
	\section{Background}
	
	\subsection{Framework of Self-Learning Search Engine}
	\noindent\textbf{Self-Learning Search Engine}. A \textit{Self-Learning Search Engine (SLSE)} is a multimedia search engine that continuously learns and evolves to adapt its answer lists to queries submitted by users, so as to provide search results according to the current user preferences while considering future utility. When a user submits a query $Q$, the search engine takes a hybrid evolutionary exploration strategy to construct and present a result retrieval list of $M$ objects \textit{M\textunderscore List} to the user for evaluation. It functions through an online learning process by building indexes to encapsulate knowledge learned from previous interactions with users, i.e. rewards received from users. Designed to overcome the difficulty in semantic multimedia search, the combination of RL with dynamic indexing helps to eliminate individual bias of users and random deviations caused by heedless errors to assure its robustness. 
	
	\noindent\textbf{Dynamic Indexing}. Dynamic indexing is an indexing technique that dynamically builds semantic indexes to associate query terms with multimedia objects. Unlike all other indexing structures that only concerns with the construction of indexes, in dynamic indexing, new query terms could be constructed as indexes for desired objects, while existing indexes are able to be deconstructed in accordance with actual demand. The relevance score of an index then continuously changes during the process. Such mechanism is achieved automatically in the course of evolution without human invention, which confers significant practical value. In order to include the richest and the latest - potentially realtime, information and to provide the most satisfying search results, search engines generally collect data from various data sources instead of storing massive complete data in a database. Such channels include but not limit to personal blogs, social media, news portals, music and video streaming media etc. Most of these websites and mobile applications require users to register accounts in order to trace user-specific information for precision marketing, including personal information, published articles and songs, personal images and other sensitive information. As more and more social dilemmas and uneasiness are caused by making personal data public online, the function that allows users to set their own privacy levels for different content is provided to address the increasing concern of privacy. When previously public multimedia information is deleted, or the level of privacy is raised to be private or group-oriented, it is obligatory for search engines to deconstruct related indexes, if any, to preserve the rights of owners. Meanwhile, illegal and misleading information, such as news that is proved to be inauthentic or infringing the rights of reputation, images that are corrected for wrong classification, and music without copyright authorization should not be promulgated with original indexes as well. Compared with manual monitoring that is time-delayed and labor-intensive, dynamic indexing automates the monitoring process for effective information dissemination and economizes on manpower. The deconstruction of indexes also helps address the newly-emerged concerns such as privacy legitimacy, and potential issues in online social networking.
	
	\noindent\textbf{Relevance Index Value}. By capturing, analyzing, and translating the selection behaviors of the users, the computation engine of SLSE creates a learning function $\mathcal{L}$ between the object space and the query $Q$ to measure the relevance liaison:
	\begin{displaymath}
	\mathcal{L}: T \times O \rightarrow \mathbb{R}^+ \cup {0},
	\end{displaymath}
	where $T$ is the space of query terms that represents the input query $Q$, and $O$ is the object space. The output of the function is the set of non-negative real numbers that specify the corresponding relevance with 0 indicating complete irrelevance. For each link between an index term $\tau$ and a multimedia object $o$, the function $\mathcal{L}$ is of the form:
	\begin{displaymath}
	\mathcal{L}(\tau, o) \rightarrow \omega\mathcal{R}, \omega \in (0, 1),
	\end{displaymath}
	where $\mathcal{R}$ is a relevance base adopted by the search engine that could be flexibly adjusted, and $\omega$ is a normalized weight that measures the degree of relevance. A larger value of $\mathcal{R}$ will increase the dispersion of indexes in the system, and vice versa. For brevity of notation, $r$ is adopted to substitute for $\omega\mathcal{R}$, which can be seen as a \textit{Relevance Index Value (RIV)} that assesses the relevance and importance of an object to a term. A higher numerical value of $r$ indicates multiple repeated learning reinforcements and hence a greater degree of importance of an object $o$ associated with a term $\tau$, and vice versa. At the beginning, all RIV scores are initialized by the system. Later in the usage, the learning function $\mathcal{L}$ takes the results of the reward function as input to update pertinent RIV scores iteratively.  Unlike extracting keywords from text-based documents, the indexing of multimedia objects is often an evolutionary learning process, since it is generally impossible to incorporate all the properties of an object in an index at the very beginning. The usage of dynamic indexing enables SLSE architecture to encapsulate the deep knowledge from higher-level human perception for effective retrieval, which is achieved by learning proactively from the relevance feedback provided by users.
	
	Here, we assume the only form of knowledge that the search engine learns from users is the implicit clicking information. When evaluating a returned \textit{M\textunderscore List}, the user is assumed to click on an object if and only if it is pertinent or not click at all if the whole list is irrelevant. For learning purpose, scalar positive or negative reward is then calculated by a reward function to adjust the future actions of the search engine, so as to provide more satisfying results to users. Random errors could be eliminated in the iterative learning process. Such clicking information represents human preferences in different stages and possesses the inherent advantage of requiring no supplementary toolkits for the collection of data compared with browser cache, logs and metadata files. Therefore, it could be readily exploited to boost up the preciseness and relevance of search results along with time.
	
	\section{A SLSE Architecture With Reinforcement Learning}
	\subsection{Reinforcement-Learning Based Architecture}
	The fact that RL intrinsically matches with the mechanism of SLSE architecture makes it a promising solution to multimedia search. Different from supervised learning with predefined input-output pairs, there is no instructive information to explicitly define correct actions for SLSE. Instead, the search engine learns and chooses an action from the action space in each round of searching based on its current state and the knowledge it possesses. Reward is then observed via the clicking information generated by the user, helping the search engine to evolve into a new state by transforming and updating the internal index pool. While ultimately aiming to retrieve the most relevant objects, the search engine architecture maintains a balance between exploitation and exploration to prevent local optima during evolution. To do this, an evolutionary exploration strategy is applied to allow the appearance of new objects while considering the relevance relationship revealed in the index learning process. 
	
	In our framework, the SLSE architecture acts as an agent that learns to act, and the user submitting query requests takes the role of the environment. In particular, the SLSE architecture is formulated as a Markov Decision Process with five-tuple \{$\mathcal{S}$, $\mathcal{A}$, $\mathcal{T}$, $\mathcal{R}$, $\gamma$\} \cite{sutton2018reinforcement}. In particular, we consider discrete states and discrete actions in continuous time, with infinite time horizon, and let $\mathcal{S}$ denote the state space, $\mathcal{A}$ denote the action space. At a specific time step $t$, the SLSE agent is in state $\mathcal{S}_t$ $\in \mathcal{S}$, choose an action \textit{a\textsubscript{t}} $\in \mathcal{A}(\mathcal{S}_t)$ to take from $\mathcal{A}$ according to the policy $\pi$. When the policy is deterministic, it is a mapping from states to actions $\pi: \mathcal{S} \rightarrow \mathcal{A}$. On the other hand, when stochastic policy is considered, $\pi$ can be seen as a state dependent \textit{probability mass function (pmf)} such that  $\pi$: $\mathcal{S} \rightarrow$ \textit{p} $(\mathcal{A}) \subset \lbrack0, 1\rbrack$ is a mapping from states to the probabilities over possible actions. In such a case, $\pi(a|s)$ gives the probability of taking action \textit{a\textsubscript{t}} = \textit{a} in state \textit{s\textsubscript{t}} = \textit{s}: 
	\begin{displaymath}
	\pi(a|s) = Pr[a_t = a | s_t = s],   s.t. \sum_{a\in\mathcal{A}} \pi(a|s) = 1.
	\end{displaymath}
	Due to the non-deterministic elements presented in the environment, taking an action \textit{a\textsubscript{t}} leads the agent to transit from the current state \textit{s\textsubscript{t}} to the next state \textit{s\textsubscript{t+1}} according to the transition kernel $\mathcal{T}$. When an observation $\mathcal{O}_t$ is made, a one-time reward signal \textit{r\textsubscript{t}} is observed given by the reward function $\mathcal{R}$(\textit{s', s, a}), which is a random variable with the following expectation: 
	\begin{displaymath}
	\mathcal{R}(\textit{s', s, a}) = \mathbb{E}[r_t|s_{t+1} = s', s_t = s, a_t = a]
	\end{displaymath}
	In each learning episode, it is essential to obtain the expected value of discounted rewards for a given policy, where $\gamma \in (0, 1)$ is the discount factor. The goal is to find optimal policies that maximize the expected infinite-horizon discounted sum of rewards $\sum_{t=0}^{\infty}\gamma^tr_t$. The corresponding long-term mission of SLSE therefore is to build up indexes to return the most relevant objects as search results to users.
	
	\noindent\textbf{Action Space}. The action space $A$ consists of a series of actions that the agent selects $M$ objects to form a \textit{M\textunderscore List} and presents it to the user. The size of the initial action space is intuitively huge and fundamentally increases the computational complexity, as it involves a huge number of possible permutations of $M$ objects. Nonetheless, our hybrid evolutionary exploration strategy will significantly decrease the size of the action space by excluding those permutations that lead to poor performance. A large number of searchable objects tend to be irrelevant to a particular query, and including them as candidate members of the permutation in each round of learning will be inefficient. We will examine further the action strategies below, but for the moment, it suffices to regard the action as selecting a certain number of objects for presentation to users in response to their query.
	
	\noindent\textbf{State Space}. The state space $S$ can be seen as an infinitely large bounded set. It is a set of all indexes in the dynamic indexing component - including both the explored and unexplored ones - together with their RIVs. In the case of unexplored indexes, their RIV are below a pre-defined threshold $h > 0$, while for explored indexes, their RIV are at or above it. After taking a particular action, RIV scores change accordingly, causing a corresponding state transition in the system. The long-term goal of SLSE is to expose unexplored indexes for satisfactory retrieval, and reward will be the net change in the total RIV scores as a result of an action. 
	
	\subsection{Problem of Learning Convergence}
	Here, a TOR-tuple [$\tau_i, o_j, r_{ij}$] can be used to represent the relationship between a query term $\tau_i$ and an object $o_j$, where $r_{ij}$ is the output calculated by the learning function $\mathcal{L}$ of the computation engine. For such a tuple to be an index, $r_{ij}$ must attain or surpass the pre-defined threshold value $h$, which indicates $o_j$ is sufficiently relevant to $\tau_i$. Assume there are two categories of indexes: explored indexes and unexplored indexes. All TOR-tuples initially reside in the index generator. When $r_{ij}$ exceeds or equal to the threshold, a TOR-tuple becomes an explored index and is squeezed out from index generator into another component named the index pool. On the other hand, an unexplored index is a potential index where oj is intrinsically sufficiently relevant to $\tau_i$, but is still in the evolutionary stage. Ideally, all unexplored indexes shall become explored indexes when the learning process is completed so that the search engine is able to function most efficiently. Note that although unexplored indexes are in the index generator before becoming explored indexes, not all TOR-tuples in the generator are unexplored indexes. Fig. 1 shows the whole framework of SLSE with RL.
	
	\begin{figure}
		\centering
		\includegraphics[width=.48\textwidth]{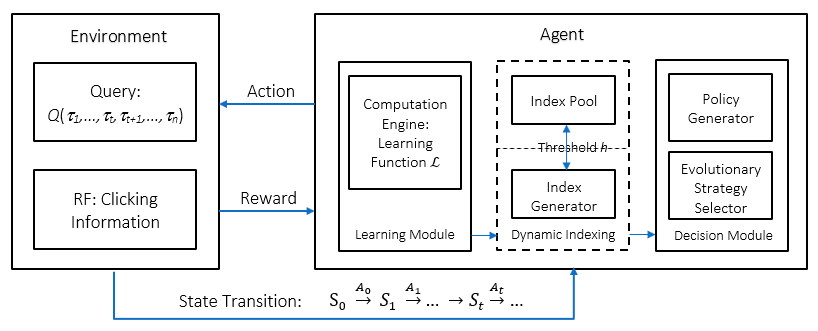}
		\caption{Framework of SLSE Architecture}
		\label{fig}
		\vspace*{-6mm}
	\end{figure}
	
	The problem of \textit{Learning Convergence} is concerned with whether unexplored indexes will eventually evolve to become explored indexes. The index learning behavior is considered to be convergent if the vast majority of unexplored indexes become explored indexes in the learning process. As objects are only traceable through query terms, each object is at least minimally indexed in the inception phase to ensure it is searchable, and a predefined RIV score $R_{init}$ is assigned for such purpose. SLSE gradually tunes the initial results to be optimal answers by learning from the observed rewards. We therefore assume that for an explored index to exist between an object oj and a term $\tau_i$, we have $r_{ij} \gg R_{init}$. When $r_{ij}$ approximately equals to $R_{init}$, the TOR-tuple [$\tau_i, o_j, r_{ij}$] fails to be installed as an explored index. That is, $o_j$ is not sufficiently relevant to $\tau_i$.  
	
	The successful construction of explored indexes guarantees appropriate objects can be effectively retrieved with semantic indexes for intelligent query answering. Notice that semantic indexes are built in an evolutionary process. Consider a query with multiple terms $Q(\tau_1,..., \tau_t, \tau_{t+1}, ..., \tau_n)$, where explored indexes exist for $\{\tau_1, ..., \tau_t\}$ with a set of multimedia objects, and $\{\tau_{t+1}, ..., \tau_n\}$ are entirely new terms input for the first time. In such case, the candidate returned objects for \textit{M\textunderscore List} of $Q$ is a union set decided by the evolutionary exploration strategies, which encompasses a $k-$object subset $O_a$: $\{o_1, o_2, ..., o_k\}$ that has the highest cumulative RIV scores, and a subset $O_b$ of several random objects selected for exploration. The algorithm of constructing $O_a$ is defined as follows:
	
	\begin{algorithm}
		\caption{Constructing $k-$object subset $O_a$}
		\begin{algorithmic}[1]
			\Require $R_{init}$, $M$, $K$, $query$ $Q(\tau_1,..., \tau_t, \tau_{t+1}, ..., \tau_n)$
			\State Initialize RIV score buffer $B$ as $\emptyset$ 
			\State Initialize $k-$object subset $O_a$ as $\emptyset$ 
			\State Initialize object counter $k$ as 0
			\For {$j = 1, ..., \vert O \vert$}
			\State Initialize $r_{Tj} \gets 0 $
			\For {$i = 1, ..., t$}
			\State Retrieve RIV score: $r_{Tj} \gets r_{Tj} + r_{ij}$
			\EndFor
			\State Append RIV score $r_{Tj}$ to buffer $B \gets B \cup \{r_{Tj}\}$
			\EndFor
			\While {$k < K$}
			\State $j = argmax_j B$  \Comment Find the index of object $o_j$ with highest $r_{Tj}$ from buffer $B$
			\State Retrieve $o_j$ and append to $O_a$: $O_a \gets O_a \cup \{o_j\}$ 
			\State Remove $r_{Tj}$ from buffer $B$
			\EndWhile			
		\end{algorithmic}
	\end{algorithm}
	
	When the final \textit{M\textunderscore List} is presented, objects of the greatest interest are expected to receive positive rewards from the user and RIV scores related to all query terms in $Q$ are boosted. After some time of evolution, explored indexes for $\{\tau_{t+1}, ..., \tau_n\}$ will be successfully constructed and the reinforcement of iterative learning process will dynamically expose the most desired objects to queries that contain any of these terms.
	
	\subsection{Evolutionary Exploration Strategies}
	The action in response to a query $Q$ consists of two steps: 
	
	(a) a selection of a set of $M$ objects, 
	
	(b) ranking of the objects for presentation. 
	
	\noindent We shall refer to this set as $M_Q$. As we indicated above, the set $M_Q$ is a union of the two sets,
	\begin{displaymath}
	M_Q = O_a \cup O_b .
	\end{displaymath}
	We shall normalize $\beta$ with $0 \leq \beta \leq 1$ to represent the proportion of objects selected based on their RIV, i.e.
	\begin{displaymath}
	\beta = \frac{\vert O_a \vert}{\vert O_a \vert + \vert O_b \vert}.
	\end{displaymath}
	
	In extreme cases where $\vert O_a \vert = 0$ or $\vert O_b \vert = 0$, then we have $\beta = 0$ or $\beta = 1$ respectively. For the time being, we let $\beta$ be deterministic and to avoid involving non-integral number of objects, we have
	\begin{displaymath}
	\beta \in \Big\{\frac{1}{\vert M_Q \vert}, \frac{2}{\vert M_Q \vert}, \dotsi, \frac{M}{\vert M_Q \vert}\Big\}.
	\end{displaymath}
	
	Thus, value for $\beta$ needs to be chosen firstly. A high value for $\beta$ risks landing on a local optimum, whereby highly relevant objects may never be presented for clicking with its RIV remaining low, while less relevant objects are repeatedly presented, leading in time to a high RIV for them. Conversely, a low value for $\beta$ may undermine retrieval efficiency, with the users often not being able to find the target objects. A variation of the above strategy is to allow $\beta$ to be random, taking values in the interval (0, 1). $\beta$ then becomes a random variable uniformly distributed over the unit interval. The number $\vert O_a \vert$ of $O_a$ objects selected is $\beta M$ rounded to the nearest integer:
	\begin{displaymath}
	O_a  = \lfloor \beta M + 1 / 2 \rfloor.
	\end{displaymath}
	
	Irrespective of whether $\beta$ is random or deterministic, we retrieve the set $O_a$ based on the algorithm given above. After this selection, let there be $J$ objects remaining in the index pool, then each of these $J$ objects will be chosen with equal probability without replacement, i.e. the first one is chosen with probability $1/J$, the second with probability $1/(J-1)$, and so on. In this way, having selected the $M_Q$ objects, step (a) is completed. Since a high-ranked object tends to receive greater attention, the set of $M_Q$ objects needs to be ordered. These may be ordered in different ways.
	
	\begin{enumerate}
		\item Completely Random Ordering
		
		Here the $M_Q$ objects are all treated the same with the top object chosen from among the $M_Q$ objects with equal probability. Similarly, the next object is chosen from among the remaining ones with equal probability. 
		
		\item Sectionally Random Ordering
		
		Here all the $O_a$ objects are all treated the same with the top object chosen from among them with equal probability similar to 1. Likewise, all the Ob objects are chosen in a similar manner. Here, the $O_a$ objects will always precede the $O_a$ objects in the ordering, i.e. $O_a > O_b$. 
		
		\item Partially Random Ordering
		
		Here we also have $O_a > O_b$. However, the $O_a$ objects are not treated the same with the top object chosen from among them based on their RIV. A particular object $o_j \in O_a$ is chosen as the top object with probability based on its relative RIV, i.e. with probability
		\begin{displaymath}
		O_a  = \frac{\sum_i r_{ij}}{\sum_{ij} r_{ij}}.
		\end{displaymath}
		Where the bottom summation is ranged over all the $O_a$ objects. Likewise, the remaining $O_a$ objects are chosen in descending order based on their relative RIV. The $O_b$ objects are chosen as in 2 above.
		
		\item Non-Random Ordering
		Here we also have $O_a > O_b$. All the $O_a$ objects are simply ranked in descending order of their RIV (i.e. the top object is the one with the highest RIV) from among the $O_a$ objects, while the $O_b$ objects are chosen as in 2 above.
		
	\end{enumerate}
	
	\begin{figure*}
		\centering
		\begin{subfigure}[t]{0.49\linewidth}
			\centering
			\includegraphics[height=3cm, width=0.49\linewidth]{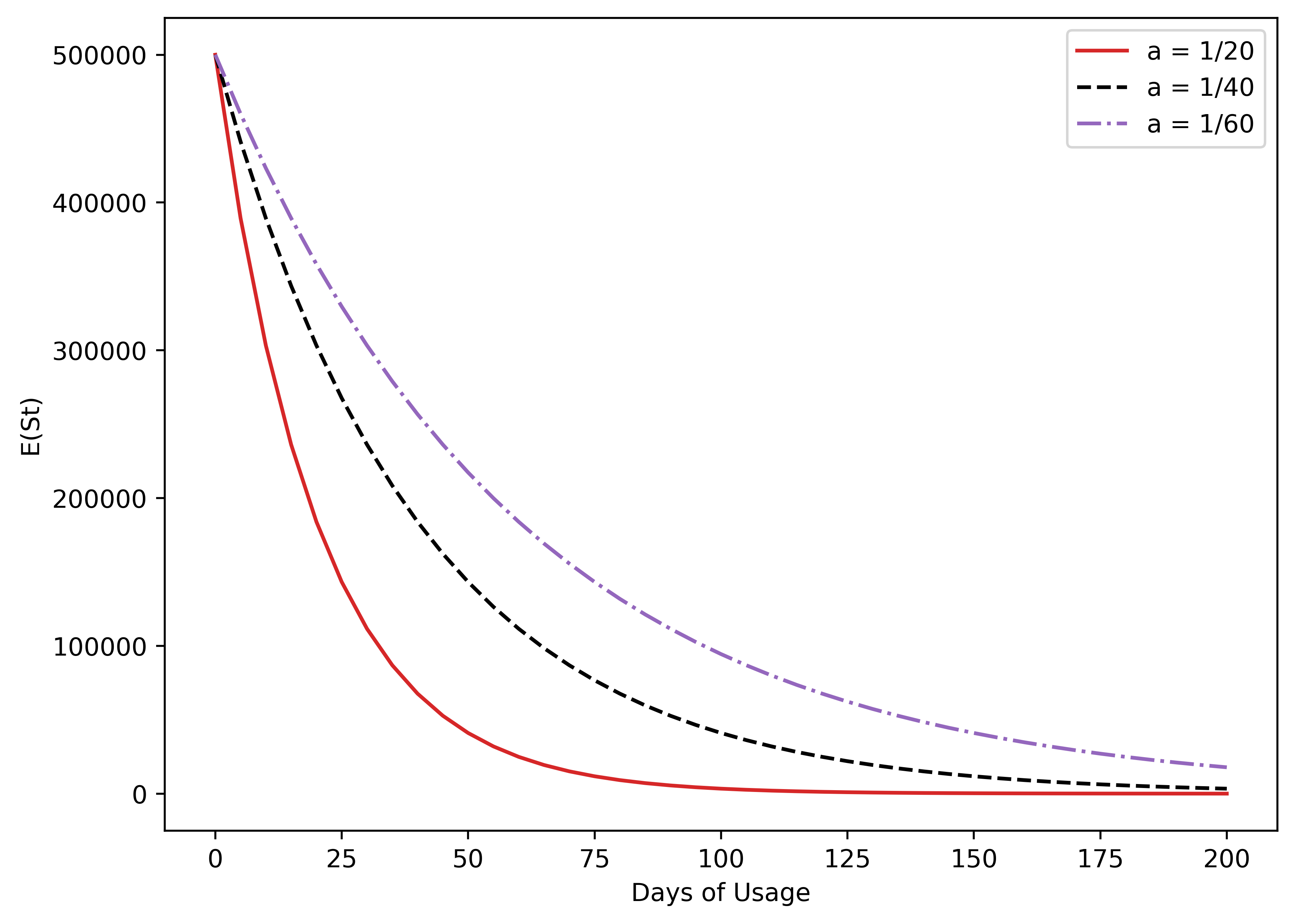}
			\includegraphics[height=3cm, width=0.49\linewidth]{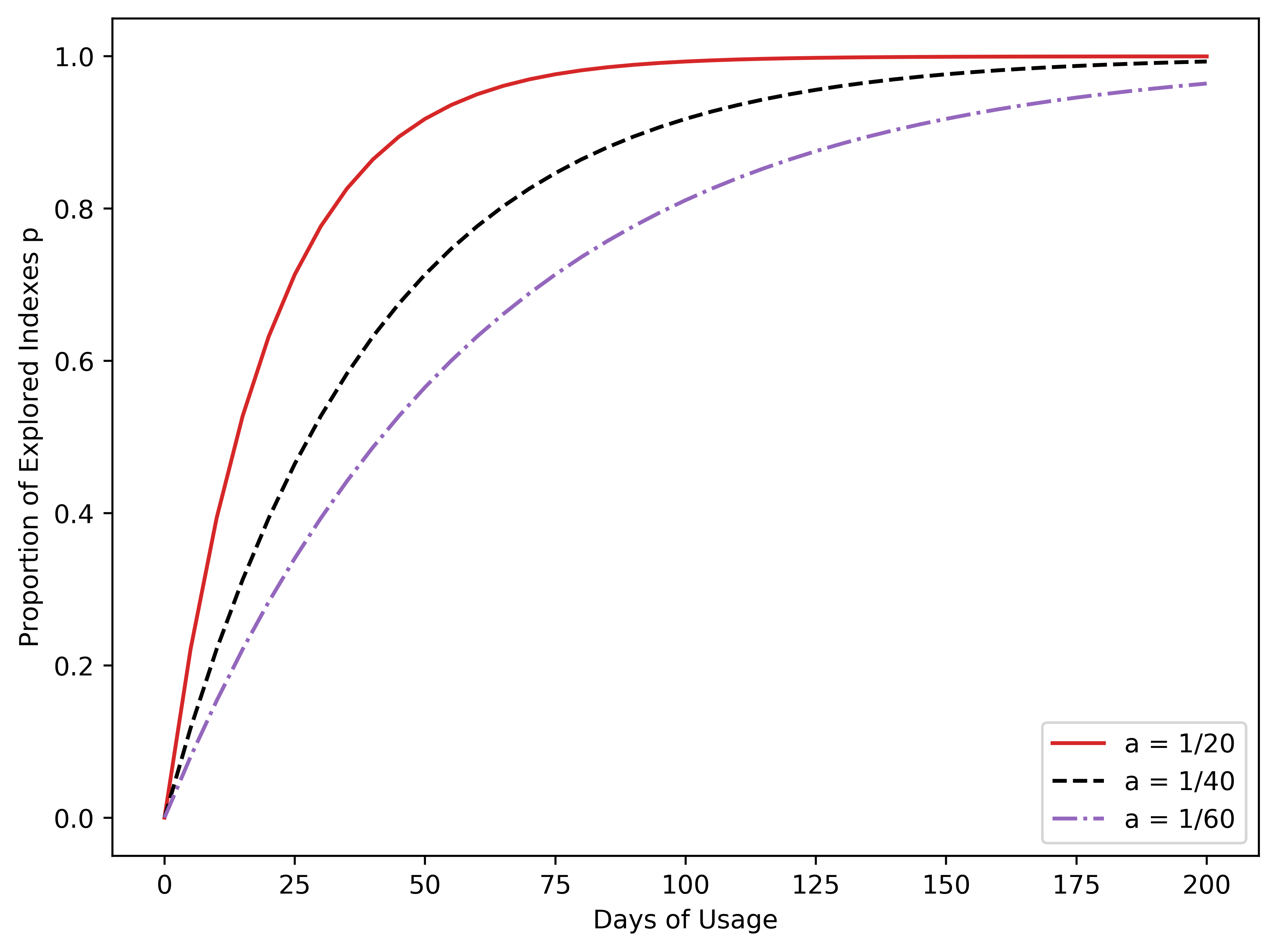}
			\caption{Behavior of $\mathbb{E}_{S_t}$(Left), $p$(Right):$S_0 = 500,000$}
		\end{subfigure}
		\begin{subfigure}[t]{0.49\linewidth}
			\centering
			\includegraphics[height=3cm, width=0.49\linewidth]{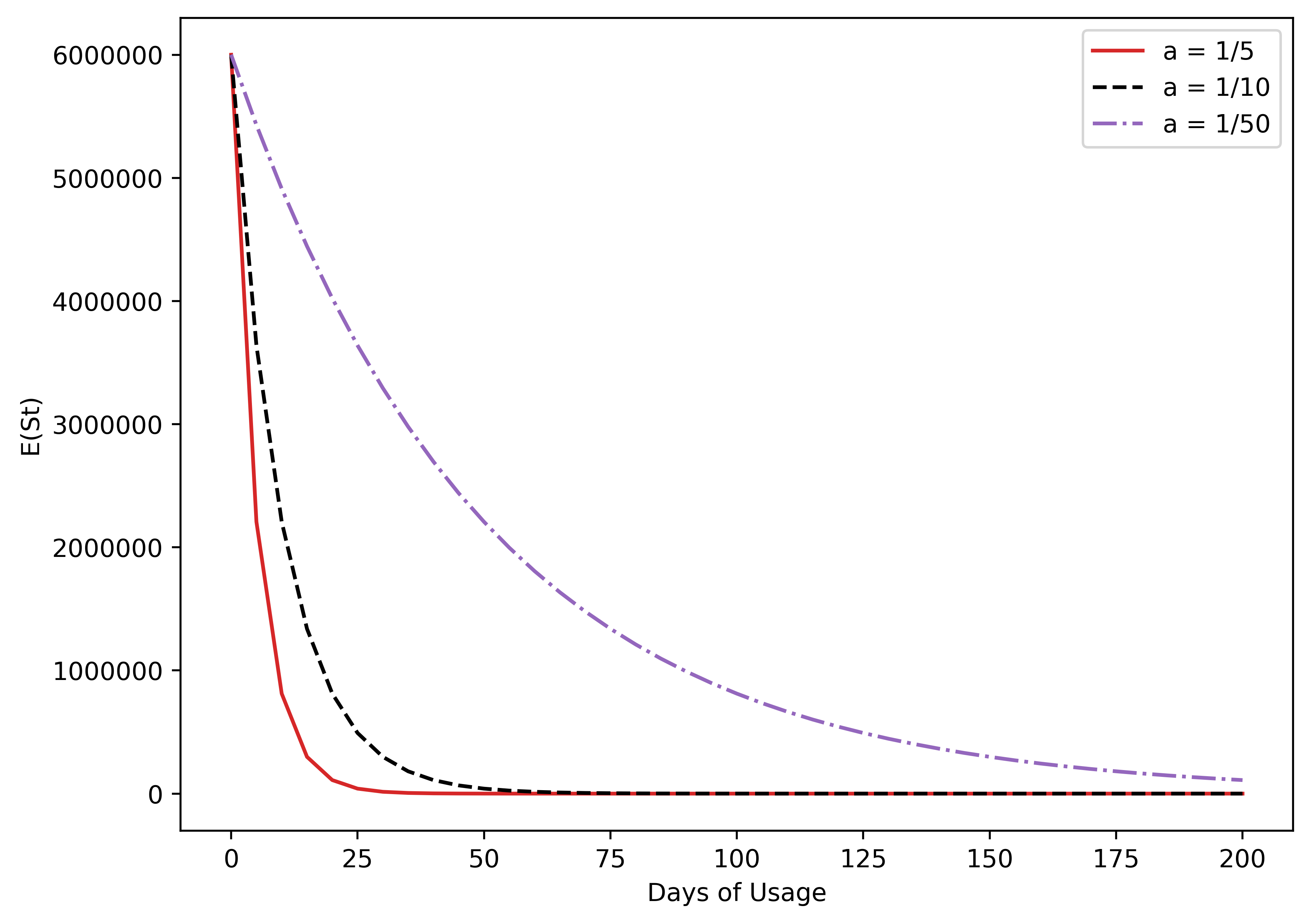}
			\includegraphics[height=3cm, width=0.49\linewidth]{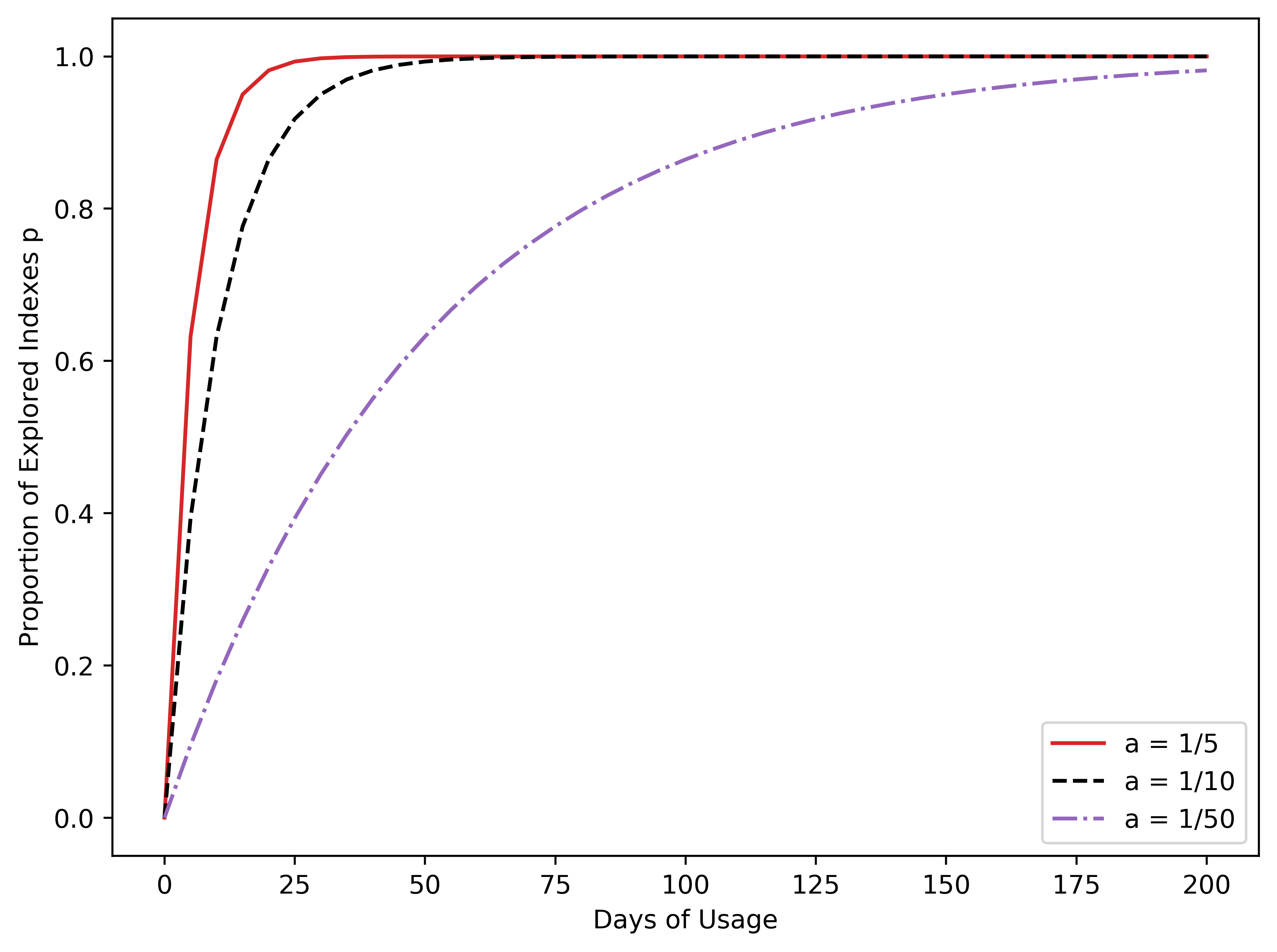}
			\caption{Behavior of $\mathbb{E}_{S_t}$(Left), $p$(Right):$S_0 = 6,000,000$}
		\end{subfigure}
		\caption{Theoretical Analysis of Convergence behavior}
		\vspace*{-6mm}
	\end{figure*}
	
	\section{Analysis of Learning Convergence Behavior}
	If an unexplored index is elevated into the index pool, it can be regarded as \textit{exposed}. Conversely, it is regarded to be extant and still in the evolutionary process. To study the characteristics of index learning behavior in SLSE, we first need to consider the evolution of a single unexplored index. Note that the time that an unexplored index evolves into an explored index tends to be random, due to the fact that the exact time point of query input is not deterministic. Let the random variable $X(t)$ be the number of times that an unexplored index is being indexed (i.e. receive reinforcement) in the time interval $(0, t)$. Since $X(t)$ are point events in time, the evolution pattern of a single TOR-tuple [$\tau_i, o_j, r_{ij}$] is a point process, and furthermore as we are dealing with a single index, we have $Pr[X(t, t+\epsilon) >1] = o(\epsilon)$. We let the random variable $dX(t)$ denote the number of times the unexplored index is indexed in the time interval $(t, t + dt)$, and we let
	\begin{displaymath}
	a(t) = \mathbb{E}[dX(t)] / dt.
	\end{displaymath}	
	The value of $a(t)$ depends on actual usage, popularity and indexing frequency of the search engine. Very often the point process is taken to be a stationary non-homogeneous process, so that $a(t) = \alpha$. In addition, for many practical situations when the point events are uncorrelated, the underlying process may be approximated by a Poisson process, in which case the probability that the unexplored index remains unindexed in the time interval $(0, t)$ is $e^{-\alpha t}$. The evolution of such an unexplored index will be considered an instantaneous failure.
	
	The counting of indexes is defined for each term-object pair. $N$ indexes are counted if a single term $\tau_i$ has explored indexes with $N$ objects, or for an object $o_j$ associated with $N$ query terms with RIV sores higher than the threshold value. Since the relevance relationship between each object and each query term is intrinsically determined under the semantic environment during a particular time frame, it is reasonable to assume there are a total of $C$ unexplored indexes in SLSE initially. The index learning process starts from time $t = 0$; for simplicity in our stochastic analysis, we shall use the reduced state space and let $S_t$ denote the reduced system state, omitting the RIV scores, so that $S_0 = C$. The utilization of rewards in RL mechanism will lead to the decrease of $S_t$ compared with $S_0$. Whenever an unexplored index is exposed, $S_t$ is decreased by one. The present situation is an instance of the general birth-death stochastic population process, and from \cite{goel2016stochastic}, the mean and variance of $S_t$ are respectively

	\begin{equation}
	\mathbb{E}(S_t) = S_0e^{-\alpha t},
	\end{equation}
	\begin{equation}
	V(S_t) = S_0e^{-\alpha t}\Big(1 - e^{-\alpha t}\Big).
	\end{equation}
	Since the limit of $\mathbb{E}(S_t) \rightarrow 0$ as $t \rightarrow \infty$, this indicates eventually, the entire collection of unexplored indexes will be fully discovered. 
	
	We also note that the limit $V(S_t) \rightarrow 0$ as $t \rightarrow \infty$, which reveals the fact that the effect of stochastic fluctuation tends to diminish in the course of the learning process, so that with the passage of time, the index learning behavior acts like a deterministic evolution, resulting in a robust and effective dynamic indexing method for multimedia search. 
	
	Let $T_s$ denote the expected time spent on indexing a proportion p of unexplored indexes. Then the proportion of exposed explored indexes during a time interval $T_s$ is:
	\begin{equation}
	p = \frac{S_0 - S_0e^{-\alpha T_s}}{S_0}.
	\end{equation}

	Fig. 2(a)(b) demonstrate the evolutionary behavior of $\mathbb{E}(S_t)$ and $p$ in the index learning process, with different parameter combinations of $S_0$ and $\alpha$ for comparison. Consider that the index learning behavior is convergent when $p$ is over 90\%. As can be seen from Fig. 2(a)(b), a higher index discovery rate $\alpha$ allows the number of extant unexplored indexes to drop more rapidly, resulting in a faster convergence of the indexing behavior within a shorter time. In Fig. 2(a), it takes the SLSE 45, 90 and 135 days to converge with the index discovery rate $\alpha$ of 1/20, 1/40 and 1/60 respectively. In particular, with the same index discovery rate but different $S_0$, we can expect the same curve shapes of $p$ as indicated by (3). Such intuition suggests that the initial number of unexplored indexes exerts no influence on the learning convergence behavior. Furthermore, a conclusion can be drawn that for a large number of unexplored indexes, higher index discovery rate plays an important role in providing satisfactory search performance, as the learning task of indexing is shared among a larger number of users, requiring less learning effort for each user and thus leading to a quick convergence. 
	
	\begin{figure*}
		\centering
		\begin{subfigure}[t]{0.75\linewidth}
			\centering
			\includegraphics[height=3cm, width=0.33\linewidth]{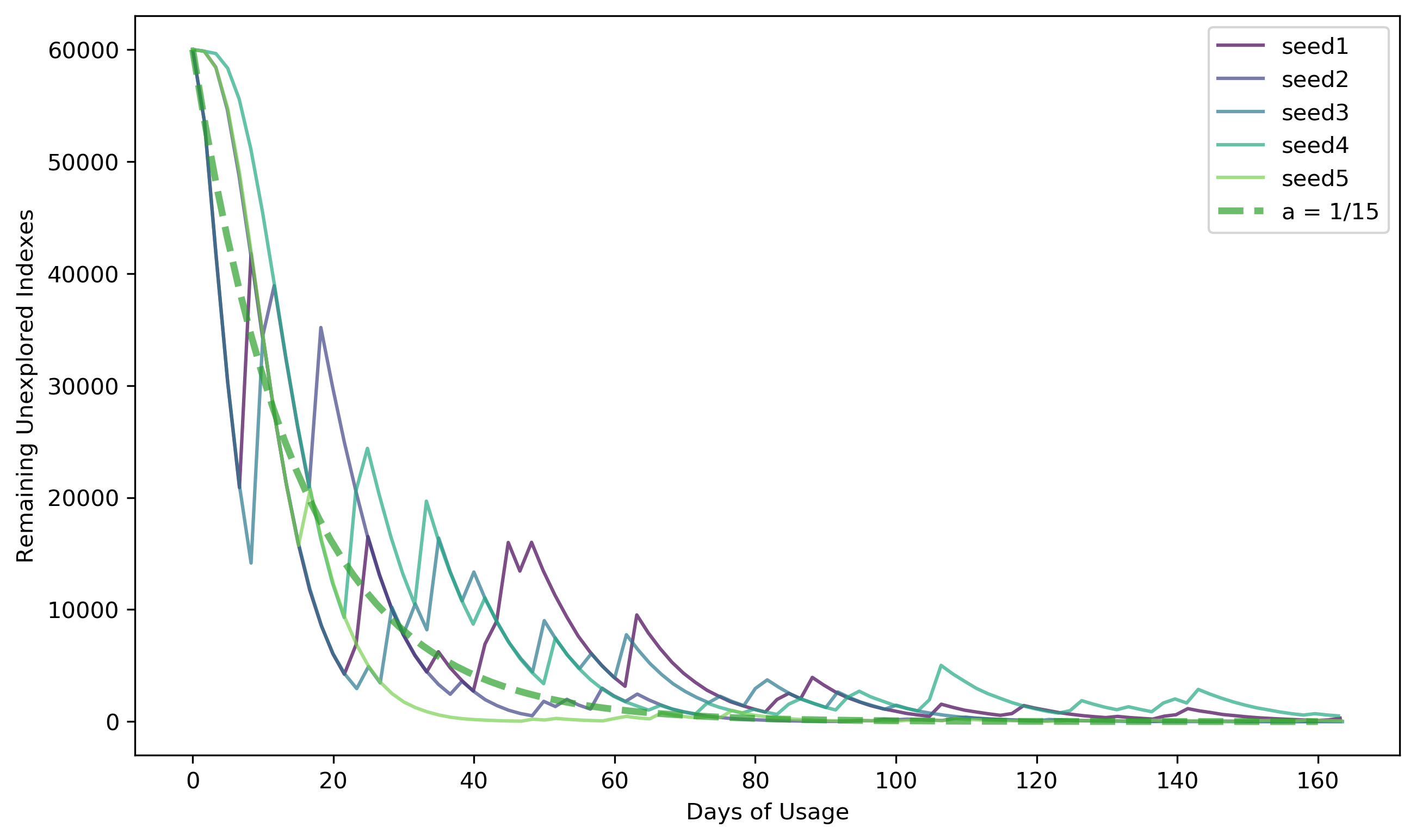}
			\includegraphics[height=3cm, width=0.33\linewidth]{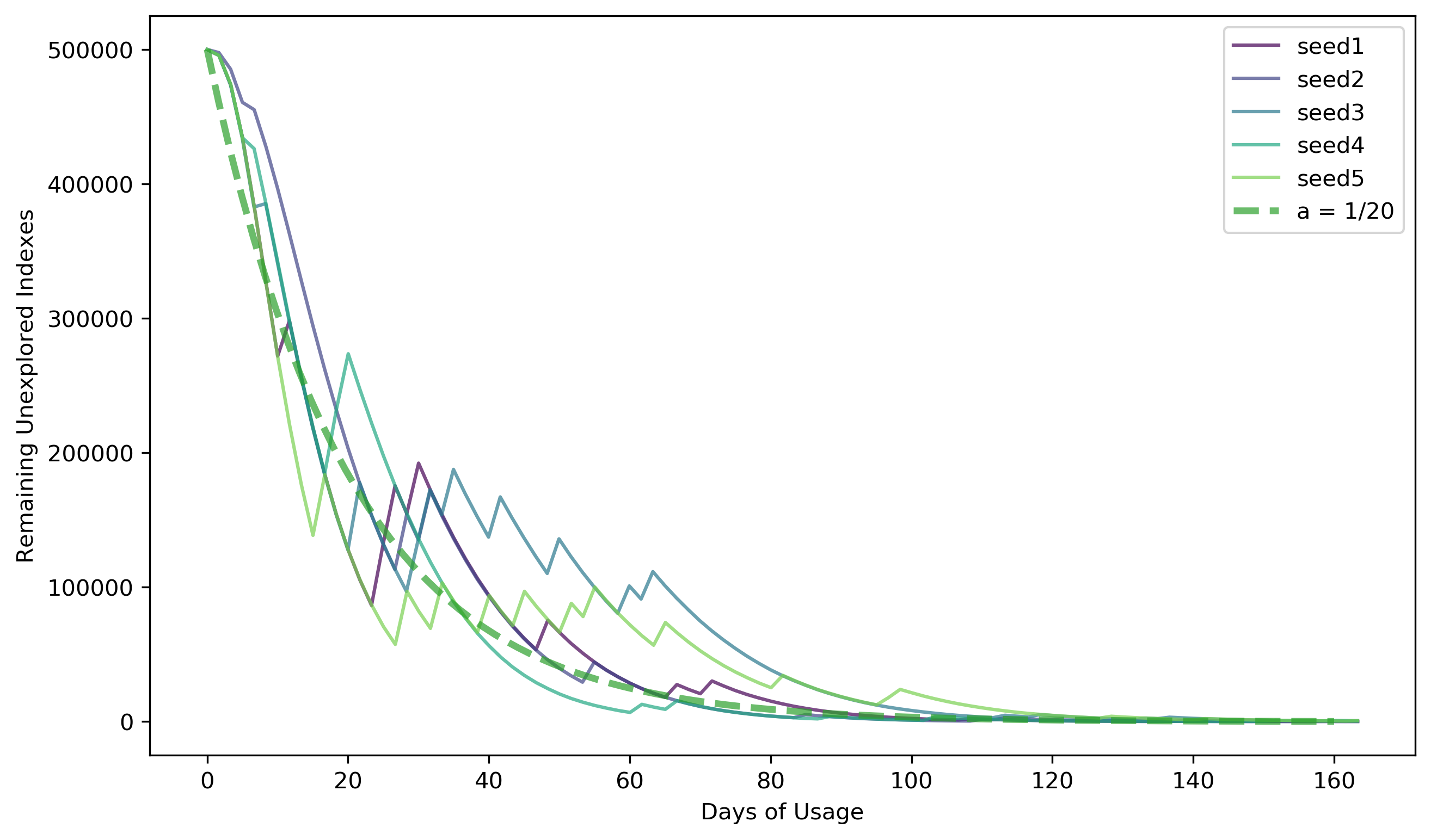}
			\includegraphics[height=3cm, width=0.32\linewidth]{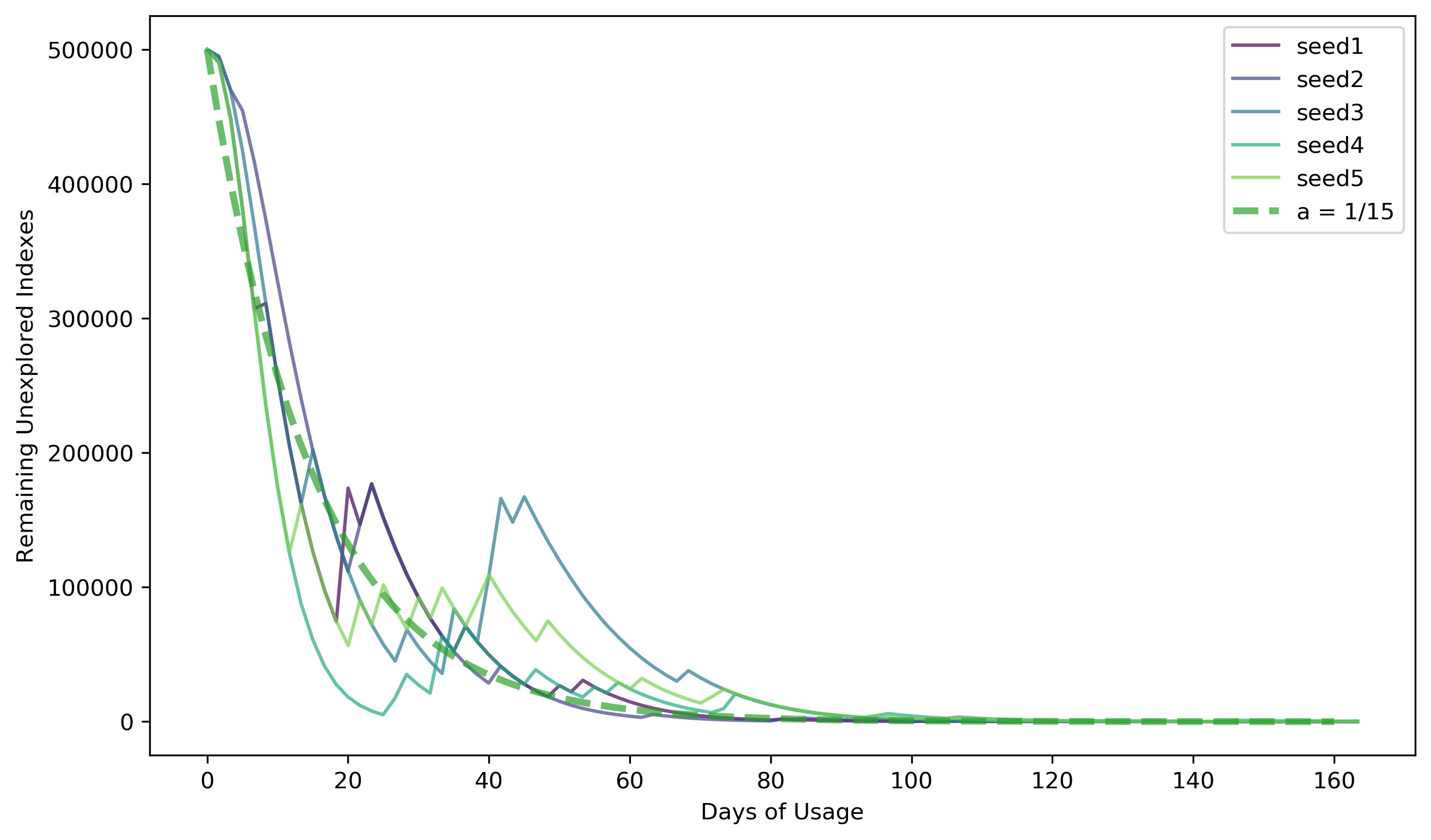}
		\end{subfigure}
		\begin{subfigure}[t]{0.24\linewidth}
			\centering
			\includegraphics[height=3cm, width=1\linewidth]{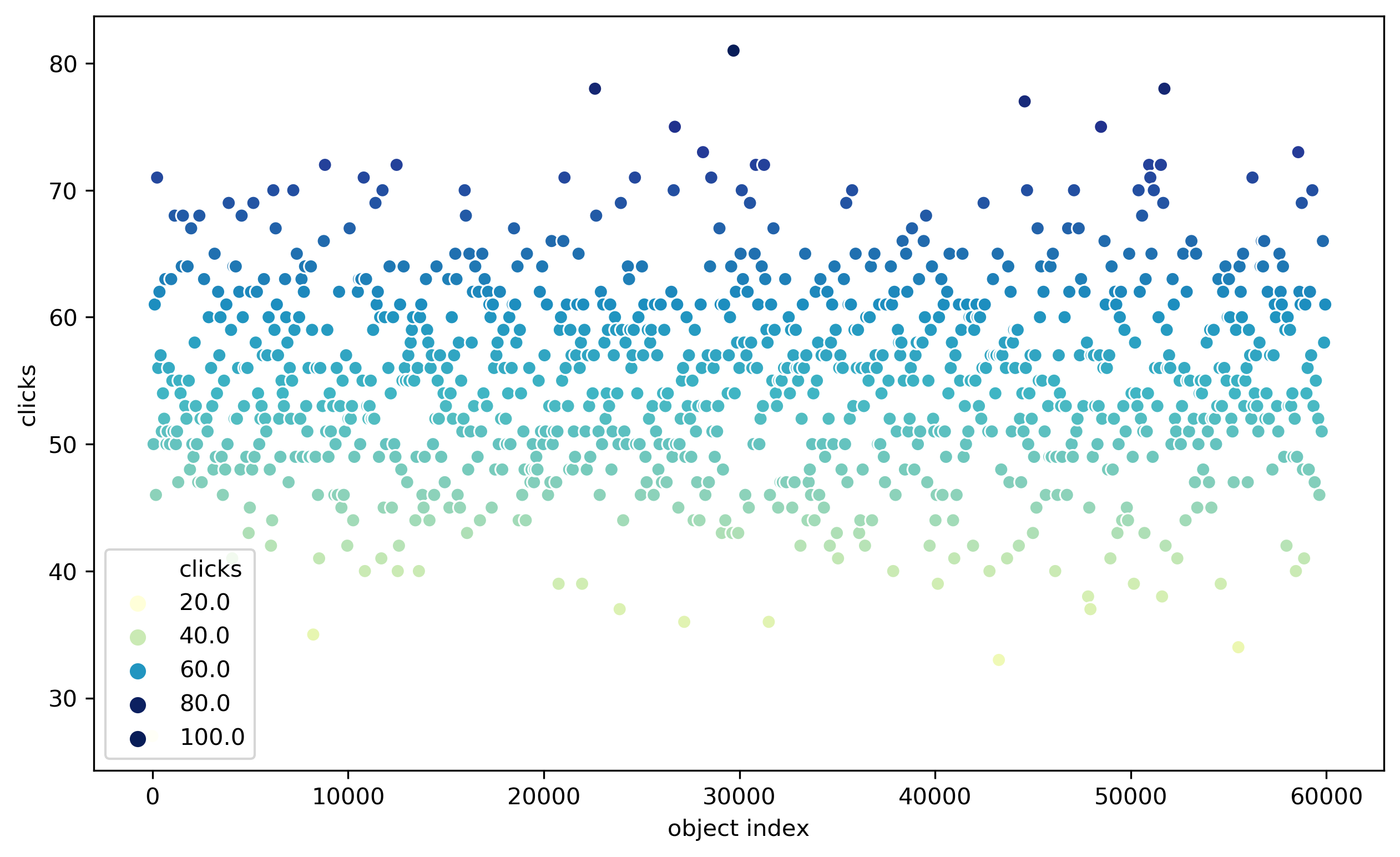}
		\end{subfigure}
		\caption{Changes of Remaining Unexplored Indexes: (a)$S_0 = 60,000; \lambda = 8,000, \alpha = 1/15$ (b)$S_0 = 500,000; \lambda = 50,000, \alpha = 1/20$ (c)$S_0 = 500,000; \lambda = 67,000, \alpha = 1/15$ (d)Overview of the Number of clicks when converged}
		\vspace*{-6mm}
	\end{figure*}
	
	\section{Experimental Evaluations}
	
	To empirically study the index learning behavior and to validate the theoretical analysis, experiments are performed to simulate the interactive learning process between users and SLSE. Specifically, we adopt Monte Carlo simulations for reproducible experimental evaluations so as to gage the validity of the results obtained above. 
	
	\subsection{Experimental Setup}
	Consider the event that an end user submits a query through the frontend interface of SLSE and furnishes evaluative feedback information by clicking interested returned objects. If such event streams are represented by a Poisson process with rate $\alpha$, then the inter-event time has the following exponential density function
	\begin{displaymath}
	f(t) = \lambda e^{-\lambda t}.
	\end{displaymath}
	When a user gains access to the frontend interface of SLSE to submit a search query, a new round of iterative learning process that leads to the internal transformation of SLSE is triggered automatically. Concerning the indexing behavior, three different scenarios exist in regard to the submitted query $Q(\tau_1,..., \tau_t, \tau_{t+1}, ..., \tau_n)$:
	\begin{itemize}
		\item \textbf{Case 1}: For each $\tau_i$ in $Q$, it is input into the system for the first time. This case typically happens in the initial stage of evolution. As introduced earlier, links of each new term are randomly established with certain multimedia objects initially, where the extent of exploration is maximized for discovery and minimal for indexing purpose. Therefore, fluctuations can vary significantly under such situation. 
		\item \textbf{Case 2}: Each $\tau_i$in Q was input into the system previously. In this scenario, the evolution of indexes for each $\tau_i$is either in progress or even completed depending on the previous iterations of learning. Results are expected to be more satisfactory compared to Case 1 and fluctuations tend to drop significantly. The relevance of each object oj to $Q$ is decided by calculating the cumulative RIV score of $o_j$ to all terms in $Q$. 
		\item \textbf{Case 3}: Query $Q$ is composed of both new and existing terms. This is a hybrid scenario that occurs continuously in the normal usage of SLSE or when significant changes of the semantic environment take place. It is more complicated compared to the above two cases in the sense that while the evolution of indexes for existing terms is still in progress, the requirement to discover indexes for new terms is raised as a new learning task. The injection of the indexes for new terms is accelerated by utilizing the previous learning outcomes for existing terms.
	\end{itemize}
	
	Note that in all the above situations, there are also two different scenarios concerning the user evaluative feedbacks, which are used to feed the reward function:
	\begin{itemize}
		\item \textbf{Case 1}: Explicit clicking information is observed for interested objects in \textit{M\textunderscore List}, resulting in the positive rewards to boost the corresponding RIV scores for enhancement in the later learning iterations. 
		\item \textbf{Case 2}: No explicit clicks are observed in the whole \textit{M\textunderscore List}, indicating the returned objects are not satisfactory as query response; SLSE will consequently castigate the objects by assigning negative rewards calculated by the reward function. 
	\end{itemize}
	
	Each time that an arrival of a user occurs, the user selection behavior is enacted by randomly clicking on some objects in \textit{M\textunderscore List}, triggering the internal changes of SLSE. To accurately assess the method, we choose multiple different random seeds to perform repetitive experiments in each case, where over 1 million queries are generated in each experiment. 
	
	\subsection{Results Interpretation}
	Fig. 3 presents graphs of the evolutionary behavior concerning the remaining number of unexplored indexes at each arrival time $t$ and the number of clicks when converged, where comparison is done with the corresponding theoretical predictions. For better visualization, we sample and retrieve the experimental results at five-day intervals. The green dotted lines represent the Monte Carlo simulation results, whereas the green lines correspond to the theoretical analysis. 
	
	It can be seen that the results of simulations match closely with the corresponding theoretical predictions with only small statistical errors. Graphs (a)(b)(c) in Fig. 3 show that in both simulation and theoretical analysis, regardless of the number of $S_0$, the indexing behavior converges as expected during the normal usage of SLSE. Because of the adoption of dynamic indexing, the number of indexes may not necessarily decrease and thus empirical results would have variance along with time. Fig. 3(b) and 3(c) also reveal that the time spent on convergence is determined by the feedback rate $\lambda$: for the same $S_0$, a greater value of $\lambda$ results in shorter convergence time. In practice, a higher feedback rate always implies the greater popularity of the search engine, as larger communities of users are inclined to use it for searching the desired information and therefore causing the learning process to evolve proactively in a shorter time. In the later stage of evolution, only few unexplored indexes remain in the index generator. It signifies a sign of the completion of the evolution, where SLSE evolves into a steady state, providing satisfactory search results to users. At the time when SLSE converges, the majority of indexes receive substantial number of clicks, signifying the effectiveness of human evaluations, as shown in Fig. 3(d).
	
	Theoretical analysis and simulation outcomes reveal the fact that a greater population of users helps provide a larger volume of useful evaluative data to guide the learning process, leading to a better search performance for SLSE. Besides correcting the perturbation caused by random initial indexing, dynamic indexing and RL techniques also ensure that the search results returned by SLSE are able to adapt to the latest requirements of the end users. It is particularly useful in a community where a group of people share similar interests.

	\section{Summary and Conclusion}
	Multimedia search is a fundamental part of effective information retrieval. Unlike text-based documents, the indexing of multimedia entities for deep knowledge retrieval is a learning and evolutionary process, since it is generally impossible to discover and index all properties of an object at a single point in time. In the current big data climate, the task to manually index online multimedia information is time-consuming, failing to keep up with the exploding amount of newly generated data, and therefore necessitating a novel automated method. By applying reinforcement learning to SLSE within a Markov decision process framework, the subtle nuances of human perceptions and deep knowledge are captured and learned for evolution, where the degree of reinforcement may be flexibly adjusted, and semantic indexes are built dynamically to interconnect search terms with the most relevant media entities. The dynamic indexing of query terms enables effective search for multimedia objects, introducing steady improvements in search performance over time.
	
	The convergence of the index learning behavior is one of the critical factors to ensure that the architectures using indexing techniques are capable to operate effectively and robustly. We employ stochastic analysis to study the evolution of index learning behavior, the results of which are able to corroborate the effectiveness of the underlying SLSE architecture. Apart from demonstrating that the vast majority of hidden unexplored indexes would be exposed during evolution, we also examine the dynamic changes of index behavior from distinct perspectives. Our study is able to show that learning convergence will be eventually achieved in the course of normal usage, indicating the SLSE architecture based on RL and dynamic indexing is both effective and efficient, conferring distinct advantages compared with conventional approaches.

	\section*{Acknowledgment}
	This research is supported by Shenzhen Fundamental Research Fund under grants No. KQTD2015033114415450 and No. ZDSYS201707251409055.

	%

\end{document}